\documentclass[11pt]{article}
\usepackage[final]{acl} % this is the new style
\usepackage{amsmath}
\usepackage{times}
\usepackage{latexsym}
\usepackage[T1]{fontenc}
\usepackage[utf8]{inputenc}
\usepackage{tablefootnote}
\usepackage{microtype}
\usepackage{booktabs}
\usepackage{inconsolata}
\usepackage{multirow}
\usepackage{graphicx}
\usepackage{amssymb}
\usepackage{pifont}

\usepackage{tcolorbox}

\title{Label-Consistent Data Generation for Aspect-Based Sentiment Analysis Using LLM Agents}

\author{Mohammad H.A. Monfared\textsuperscript{$\clubsuit$} \hspace{3mm} Lucie Flek\textsuperscript{$\clubsuit\spadesuit$} \hspace{3mm} Akbar Karimi\textsuperscript{$\clubsuit\spadesuit$}\\
\vspace{-1mm}\\
\textsuperscript{$\clubsuit$}Bonn-Aachen International Center for Information Technology, University of Bonn, Germany\\
\textsuperscript{$\spadesuit$}Lamarr Institute for Machine Learning and Artificial Intelligence, Germany\\
\texttt{ak@bit.uni-bonn.de}}

\begin{document}

\maketitle
\begin{abstract}
We propose an agentic data augmentation method for Aspect-Based Sentiment Analysis (ABSA) that uses iterative generation and verification to produce high-quality synthetic training examples. To isolate the effect of agentic structure, we also develop a closely matched prompting-based baseline using the same model and instructions. Both methods are evaluated across three ABSA subtasks—Aspect Term Extraction (ATE), Aspect Sentiment Classification (ATSC), and Aspect Sentiment Pair Extraction (ASPE)—four SemEval datasets, and two encoder–decoder models: \texttt{T5-Base} and \texttt{Tk-Instruct}. Our results show that the agentic augmentation outperforms raw prompting in label preservation of the augmented data, especially when the tasks require aspect term generation. In addition, when combined with real data, agentic augmentation provides higher gains, consistently outperforming prompting-based generation. These benefits are most pronounced for \texttt{T5-Base}, while the more heavily pretrained \texttt{Tk-Instruct} exhibits smaller improvements. As a result, augmented data helps \texttt{T5-Base} achieve comparable performance with its counterpart.
\end{abstract}

\section{Introduction}

Aspect-Based Sentiment Analysis (ABSA) seeks to identify aspect terms in text and determine the sentiment expressed toward each one \cite{SemEval14, SemEval15, SemEval16}. Despite its value in applications such as customer feedback analysis, ABSA remains difficult because it requires fine-grained, aspect-level annotations that are costly to produce and limited in coverage. As a result, models often struggle with rare aspect–sentiment combinations and domain-specific linguistic variation. Recent work has explored synthetic data generation using large language models \cite{ITRD, SelfCorrection}, but most approaches rely on prompting, which frequently produces label inconsistencies, limited structural diversity, and insufficient handling of corner cases. 

In this work, we investigate whether agentic, multi-step data generation can address these limitations. We develop an augmentation pipeline that uses Qwen2.5 \cite{bai2023qwen} within a ReAct-style \cite{yao2022react} framework, separating generation and evaluation into dedicated agents equipped with tools for style extraction, policy construction, and label verification.
The novelty of this approach lies in its task specific decomposition rather than a feedback loop. This structured approach ensures that synthetic examples strictly adhere to the complex requirements of aspect-based tasks. 
To isolate the effect of structured generation, we compare this system against a prompting baseline that uses the same model, prompts, and sampling strategy.

\begin{figure}[t]
    \centering
    \includegraphics[width=0.46\textwidth]{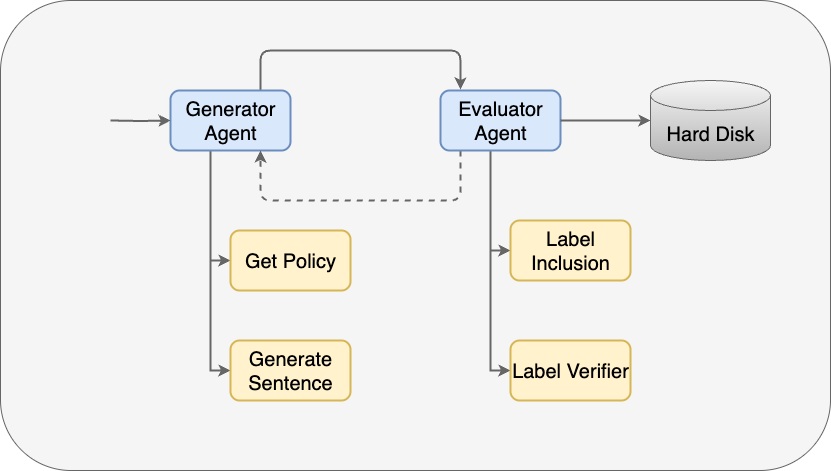}
    \caption{Overview of the agentic data augmentation workflow. A generator agent first extracts a style policy and produces candidate sentences, which are then evaluated by an evaluator agent. Only validated examples are saved, forming a high-quality synthetic dataset.}
    \label{fig:agent_workflow}
\end{figure}

Empirically, we show that, compared to the raw prompting method, the agentic pipeline generates data, particularly those containing aspect terms, whose labels are more consistent with those of the original data. Furthermore, while the resulting synthetic data alone cannot replace human annotations, the agentic augmentation—when mixed with real data—improves the performance for \texttt{T5-Base} and consistently outperforms naive prompting. We also find that augmentation benefits depend on the ABSA subtask and the underlying model: A simpler task (ATSC) and a less instruction-tuned architecture (\texttt{T5-Base}) gain the most, while the heavily pretrained model (\texttt{Tk-Instruct}) shows smaller improvements. Our contributions are threefold: 1) We introduce an agentic workflow for generating high-quality, task-specific synthetic data for ABSA; 2) We design a matched prompting baseline to isolate the contribution of structured, self-reflective generation; 3) We provide a systematic analysis of how model pretraining, augmentation scale, and subtask complexity shape the effectiveness of synthetic data.

\section{Related Work}
\paragraph{Promptiong LLMs for Data Augmentation}
\label{sec:LLM_for_augmentation}
Generating labeled data is a time-consuming and labor-intensive task. Traditional augmentation methods either function locally \cite{hsu2021semantics, karimi2021aeda} or globally but in the embedding space, which reduces their controllability \cite{karimi2021adversarial}. 
LLMs provide an alternative to manual annotation by automatically generating labeled training data for supervised learning tasks \cite{supergen, ZEROGEN, PROGEN}. 
\citet{Meng2023LLMDataAugmenters} show that while LLMs can produce synthetic samples for text classification, the usefulness of this data varies with task characteristics, with more subjective or complex tasks showing reduced benefit from synthetic examples. \citet{SelfCorrection} study intrinsic self-correction in LLMs, where models are prompted to revise their own outputs after initial generation using only their internal capabilities, without external knowledge—highlighting the potential for autonomous quality improvement. 

Applying LLM augmentation to structured problems like ABSA requires deeper task understanding, with its subtasks relying on fine-grained, aspect-level sentiment labels \cite{SemEval14, SemEval15, SemEval16}. \citet{UDA} adapt Unsupervised Data Augmentation for ABSA by introducing an MLM-based unmasking strategy to accommodate span-level structure and preserve token alignment. \citet{SPM} propose Selective Perturbed Masking (SPM), which modifies low-importance tokens while maintaining label consistency, enhancing lexical diversity. \citet{ITRD} introduce Iterative Data Generation (IDG), where LLMs iteratively produce and filter pseudo-labeled ABSA data.

These techniques mostly rely on single-step prompting, perturbation-based generation, or sentence-level scoring using LLMs. Although some approaches incorporate LLM-based evaluation, they typically infer sentiment or quality from the generated text using heuristic or soft scoring criteria rather than explicitly checking whether the intended aspect–sentiment labels were realized as intended. Our approach instead introduces an agentic pipeline that separates sampling, style extraction, controlled generation, and automatic verification, using LLMs to ensure that generated sentences actually reflect the intended labels. This design leads to higher label consistency and structural alignment in generated ABSA data.

\paragraph{Agentic Data Augmentation}
\label{sec:agentic_DA}
Recent research investigated multi-agent and iterative generation frameworks to improve synthetic data quality. HydraGAN \cite{HydraGAN} introduces a multi-agent generative framework in which multiple discriminators jointly guide the data generation process. MAG-V \cite{MAG-V} introduces a multi-agent framework in which agents generate synthetic examples and verify them through alternate question reconstruction, enhancing the reliability of generated data. APIGen-MT \cite{Prabhakar2025APIGenMT} extends agentic generation to multi-turn settings by coordinating a two-phase pipeline that produces detailed task blueprints and uses simulated agent–human interaction to generate and verify complete multi-turn data. 

These methods explored synthetic data generation in various domains, but they were not designed for the requirements of structured sentiment analysis problems such as ABSA. They also do not incorporate ABSA-specific constraints like aspect-span correctness or polarity alignment. Our work differs by introducing an ABSA-focused agentic data augmentation strategy that uses multi-step reasoning, controlled generation, and automatic verification to produce more consistent and task-aligned synthetic ABSA data.

\section{Methodology}

\subsection{Agentic Data Augmentation}

Our augmentation system is built around a ReAct-style agent architecture in which an LLM selects and invokes tools to perform subtasks such as style extraction, sentence generation, and label verification. The workflow (Figure \ref{fig:agent_workflow}) consists of two cooperating agents—a generator and an evaluator—that operate under clearly defined roles and constraints. Importantly, both the agentic and baseline prompting methods use the same underlying model (Qwen2.5-14B) and nearly identical prompts; the only difference lies in the additional reasoning and validation steps introduced by the agentic framework. This ensures a controlled comparison of augmentation strategies.

\subsubsection{Generator agent}
The generator agent is responsible for producing candidate sentences that reflect ABSA structure and exhibit stylistic similarity to real customer reviews. It relies on two tools to accomplish this task. The first tool, \textbf{get policy}, prepares all metadata required for generation. It samples one to four aspect terms and their sentiment polarities from the the SemEval training data, ensuring realistic label distributions. To guide stylistic variety, the tool also extracts writing style, grammatical structure, and approximate sentence length from a small set of real sentences drawn from the dataset. These components are combined into a generation \texttt{policy}, which acts as a blueprint for the construction of each synthetic example.

Once the policy is formed, the generator agent invokes its second tool, \textbf{generate sentences}, which uses the metadata to produce a candidate ABSA-style sentence. The underlying language model receives a structured instruction template that enforces both stylistic guidance and strict adherence to the sampled aspect–polarity pairs. The tool returns the generated sentence followed by machine-readable \texttt{Terms=} and \texttt{Polarity=} annotations, ensuring compatibility with downstream training pipelines. After this step, the candidate sentence is passed to the evaluator agent.

\subsubsection{Evaluator Agent}

The evaluator agent ensures that only valid, label-consistent examples are added to the synthetic dataset. It also operates using two specialized tools. The first, \textbf{label inclusion}, checks whether the generated sentence contains all required aspect terms exactly as specified in the policy. This step identifies structural deviations such as pluralization changes, partial matches, or missing terms, which frequently occur in naive prompting scenarios.

If the candidate passes this check, it is forwarded to the second tool, \textbf{label verifier}, which evaluates whether each aspect term is associated with the correct sentiment polarity. The tool prompts the language model to judge the alignment between the sentence and the intended sentiment labels and returns an OK or NOT OK decision. Only examples that satisfy both structural and semantic criteria are accepted into the synthetic dataset; all others are discarded, and the generator agent is prompted to create a new example.

\begin{table*}
\centering
\begin{tabular}{|c|c|}
\hline
\textbf{Raw Prompting} & \textbf{Agentic Approach} \\
\hline
\begin{minipage}[t]{0.47\textwidth}
\scriptsize
You are a critic who can generate comments on the specified aspect and sentiment. \\
We would like you to complete a sentence generation task. Please follow these requirements: \\
- You need to use the sentiment, the aspect mentioned in the prompt \\
- Domain: Restaurants \\
- Your response must include: \\
1. The sentence. \\
2. A line that starts with \texttt{Terms=} followed by the list of aspect terms used. \\
3. A line that starts with \texttt{Polarity=} followed by the matching polarity list. \\
- ALL aspect terms must appear as actual aspects in the sentence with intended polarities \\
- Generated sentence must have the writing style and grammar structure and length of this sentence: \{sent\} \\
- The sentence should not have aspect words not specified in the prompt \\
- DO NOT repeat the input text in the output \\
- PRINT ONLY THE ANSWER TEXT — no explaining, nothing else \\
- Make sure to use aspect words in the output \\[0.5em]
Good Examples:\\
\texttt{['prices'] ['negative']} \\
The prices were too high for this type of restaurant \\
\texttt{['Gnocchi', 'cheesecake'] ['positive', 'negative']} \\
The Gnocchi was perfectly cooked and delicious, but the cheesecake was dry and flavorless. \\[0.5em]
Bad Example:\\
\texttt{['soup'], ['positive']} \\
The udon soup was rich and flavorful. (term incorrect) \\
\textbf{Make sure your output exactly follows this format.}
\end{minipage}

&
\begin{minipage}[t]{0.47\textwidth}
\scriptsize
\textbf{Generate Sentence Prompt:} \\
You are a critic who can generate comments on the specified aspect and sentiment. \\
We would like you to complete a sentence generation task. Please follow these requirements: \\
- Generate a sentence using this aspect term: \{aspect\_term\} with the following polarities: \{polarity\} \\
- Write in the style: \{writing\_style\}, and use a \{grammar\_structure\} grammatical structure and \{sentence\_length\} sentence length. \\
- Domain: Restaurants \\
- Include: \\
1. The sentence. \\
2. A line that starts with \texttt{Terms=} followed by the list of aspect terms used. \\
3. A line that starts with \texttt{Polarity=} followed by the matching polarity list. \\
- Follow the exact structure shown in the examples. \\
- Do not include explanations. \\
Use plain apostrophes (') — do not escape with backslashes. \\[0.5em]

\textbf{Label Verifier Prompt:} \\
You are an expert in linguistic evaluation. \\
Check if the given aspect terms and polarities are correct for the provided sentence. \\
- If all aspect terms appear as actual aspects in the sentence with intended polarities, respond only with: \texttt{OK} \\
- If any term is missing, incorrect, or has the wrong polarity, respond only with: \texttt{NOT\_OK} \\
Do not provide explanations or any other text. \\
Example Input: \\
The food was lousy... Terms=['food'] Polarity=['negative'] → OK \\
The udon soup was rich and flavorful. Terms=['soup'] Polarity=['positive'] → NOT\_OK
\end{minipage}
\\
\hline
\end{tabular}
\caption{Side-by-side comparison of raw prompting and agentic prompt structures. Both methods share identical task instructions and output constraints; the agentic approach differs only by decomposing generation and verification into separate steps, enabling automatic rejection of label-inconsistent outputs.}

\label{tab:prompt_compare}
\end{table*}

\subsection{Prompt-Based Data Generation}

To isolate the effect of agentic reasoning, we include a simple prompting-based baseline. This method uses the same underlying language model, the same aspect–polarity sampling procedure, and nearly the same instructions as the agentic generator, but it produces each sentence through a single monolithic prompt without any style extraction, tool usage, or verification steps. The prompting baseline therefore represents a minimal augmentation strategy against which the benefits of the agentic workflow can be directly measured. Full prompts are provided in Table \ref{tab:prompt_compare}.

\section{Experiments}

We evaluate our methods on the four standard SemEval ABSA datasets (Laptop14, Rest14, Rest15, and Rest16 \cite{SemEval14, SemEval15, SemEval16}) covering the ATE, ATSC, and ASPE subtasks. All experiments follow the InstructABSA \cite{INSTRUCTABSA} framework, using its preprocessing, training, and evaluation pipeline for comparability. We adopt the InstructABSA framework since it represents one of the highest-performing and recent benchmarks for instruction-based ABSA. Using its established pipelines ensures that our results are grounded in current standard practices and remain comparable with existing benchmarks. We fine-tune two encoder–decoder models: \texttt{T5-Base} \cite{T5} and \texttt{Tk-Instruct-base} \cite{TK-INSTRUCT}. Both models share the same architecture, but \texttt{Tk-Instruct} is extensively instruction-tuned on over 1,600 NLP tasks, unlike \texttt{T5-Base}. All hyperparameters match the InstructABSA\footnote{\url{https://github.com/kevinscaria/InstructABSA}} defaults, with the only modification being an increased fine-tuning budget of 20 epochs. Each experiment is repeated three times, and we report averaged scores for reliability.

We use an automated experimental pipeline to systematically control training configurations, including data source (original, generated, mixed), augmentation ratio, and target subtask, ensuring consistent and fair comparison across all settings. Synthetic training examples are generated with the Qwen2.5-14B model using the Ollama framework \footnote{\url{https://ollama.com}}. Both augmentation strategies—raw prompting and our proposed agentic workflow—use the same base model, the same prompts, and the same sampling procedure for aspect–polarity pairs. We evaluate three data configurations: (1) training on the original SemEval data only, (2) training on synthetic data only, and (3) training on a mixture of original and synthetic data. We Also test augmentation scales of x1 (equal to the size of the original training set) and x2 (twice the size).

\begin{table*}[t]
\centering
\small
\setlength{\tabcolsep}{6pt}
\begin{tabular}{llllccc}
\toprule
\textbf{Model} & \textbf{Data Source} & \textbf{Training Strategy} & \textbf{Ratio} & 
\textbf{ASPE} & \textbf{ATE} & \textbf{ATSC} \\
\midrule
T5-Base & Original  & Original & -- & 78.84 & 88.15 & 86.65 \\
T5-Base & Agentic   & Gen     & 1.0 & 39.16 & 54.56 & 81.72 \\
T5-Base & Agentic   & Gen     & 2.0 & 39.45 & 54.10 & 82.26 \\
T5-Base & Agentic   & Mixed   & 1.0 & \textbf{79.6} & \textbf{88.66} & \textbf{88.10} \\
T5-Base & Agentic   & Mixed   & 2.0 & 78.26 & 88.58 & 87.37 \\
T5-Base & Prompting & Gen     & 1.0 & 31.29 & 49.97 & 80.53 \\
T5-Base & Prompting & Gen     & 2.0 & 30.60 & 47.90 & 79.387 \\
T5-Base & Prompting & Mixed   & 1.0 & 78.26 & 87.90 & 87.27 \\
T5-Base & Prompting & Mixed   & 2.0 & 76.88 & 87.67 & 87.08 \\
\midrule
Tk-Instruct & Original  & Original & -- & 80.84 & 89.45 & 88.25 \\
Tk-Instruct & Agentic   & Gen     & 1.0 & 42.17 & 54.69 & 80.79 \\
Tk-Instruct & Agentic   & Gen     & 2.0 & 42.54 & 53.05 & 81.22 \\
Tk-Instruct & Agentic   & Mixed   & 1.0 & 80.31 & \textbf{89.84} & 87.81 \\
Tk-Instruct & Agentic   & Mixed   & 2.0 & 80.02 & 89.07 & 87.54 \\
Tk-Instruct & Prompting & Gen     & 1.0 & 31.98 & 46.40 & 79.34 \\
Tk-Instruct & Prompting & Gen     & 2.0 & 30.62 & 44.67 & 78.39 \\
Tk-Instruct & Prompting & Mixed   & 1.0 & 78.40 & 88.88 & 87.24 \\
Tk-Instruct & Prompting & Mixed   & 2.0 & 79.15 & 87.98 & 87.21 \\
\bottomrule
\end{tabular}
\caption{ F\textsubscript{1} scores of training strategies averaged over data sets and runs.}
\label{tab:main_results}
\end{table*}

\section{Results and Discussions}
\label{sec:Results}

Table \ref{tab:main_results} provides an overall summary of model performance across all tasks, datasets, and augmentation strategies. Several clear patterns emerge. First, training exclusively on synthetic data leads to a substantial drop in performance for both models, with prompting-based data performing worst and agentic data offering a noticeably higher baseline. Second, when synthetic data is combined with real training examples, agentic augmentation consistently improves or closely matches the original-data-only performance, particularly for \texttt{T5-Base}, which shows gains across all three ABSA subtasks. \texttt{Tk-Instruct} benefits more modestly from augmentation, reflecting its stronger instruction-tuned prior, but still shows stable or slightly improved performance with agentic data. In contrast, prompting-based augmentation is less reliable. While it occasionally matches the original baseline, it often fails to provide meaningful improvements and sometimes degrades performance, especially in the more structurally complex tasks, such as aspect sentiment pair extraction and aspect term extraction.

A further observation is that increasing the amount of synthetic data beyond a 1:1 ratio generally does not yield additional gains. For both models and both augmentation strategies, the x2 setting either plateaus or slightly reduces performance relative to x1, indicating that larger quantities of synthetic data amplify noise rather than add useful training signal. Overall, these results show that high-quality, agentically generated data can enhance ABSA performance when used in moderation, whereas naive prompting and large synthetic datasets offer limited benefit.

\subsection{Label Consistency of Generated Data}
\label{sec:label_cons}

Label accuracy is critical for synthetic ABSA data, as LLM-generated sentences may alter aspect terms or express incorrect polarities, introducing noise that harms downstream training. To measure this, we use \texttt{Tk-Instruct} model (fine-tuned on SemEval-2016 Restaurants) as an automatic judge and compute how many generated examples preserve their intended aspect–polarity labels. We evaluate 1,800 synthetic samples from both the agentic and prompting methods, matching the size and conditions of the original training set.

It is important to note that the \texttt{Tk-Instruct} model used here serves only as an independent judge and was not involved in the data generation process. All the synthetic data was produced using Qwen2.5. We chose a fine-tuned Tk-Instruct for this measurement because of its high accuracy. Its judgments are used purely for analysis and have no influence on the synthetic data used for training.

Table \ref{tab:label_cons} shows that agentic data has substantially higher label consistency across all tasks, with especially large gains in ATE and ASPE. These results confirm that the verification steps in the agentic workflow produce cleaner, more reliable synthetic labels than raw prompting.

\begin{table}
\centering
\small
\setlength{\tabcolsep}{8pt}
\begin{tabular}{lccc}
\toprule
\textbf{Data Type} & \textbf{ATE} & \textbf{ATSC} & \textbf{ASPE}\\
\midrule
Agentic & 78.17\% & 80.37\% & 33.89\%\\
Prompting & 43.89\% & 72.94\%& 18.33\%\\
\bottomrule
\end{tabular}
\caption{Label consistency of augmented data from Rest16 dataset, measured by \texttt{Tk-Instruct}. }
\label{tab:label_cons}
\end{table}

\subsection{Training on Generated Data}
\label{sec:gen}

\begin{figure}
    \centering
    \includegraphics[width=0.47\textwidth]{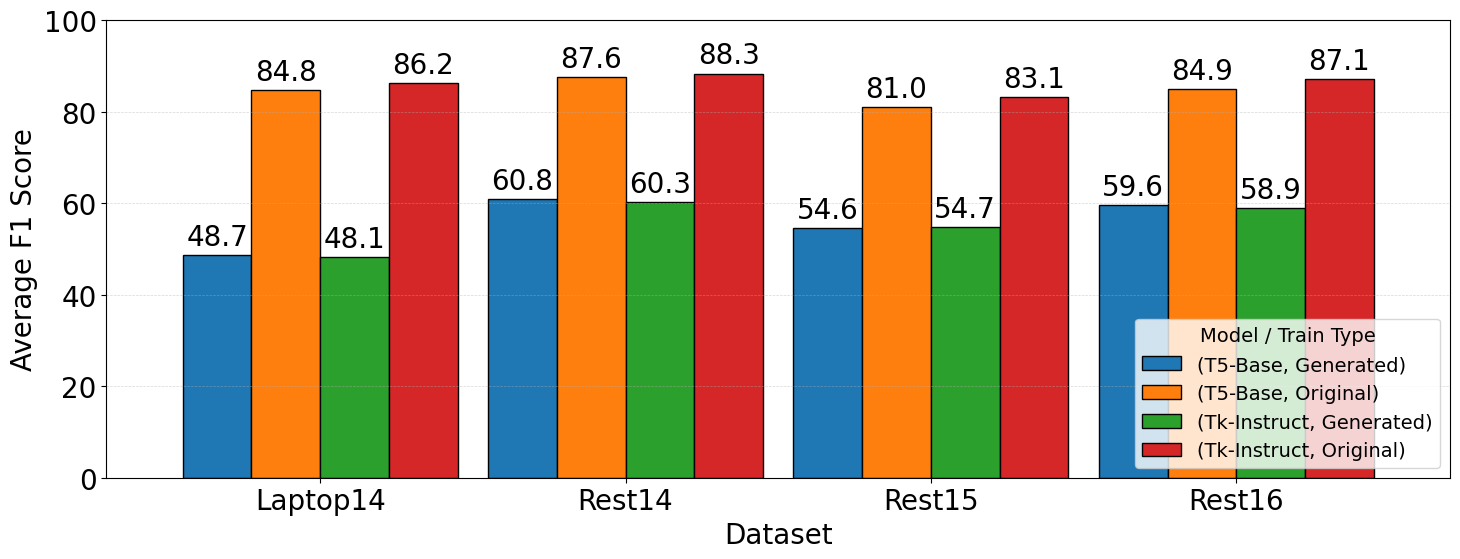}
    \caption{Average F\textsubscript{1} score across all ABSA tasks for each dataset, comparing original vs. generated-only training, signaling the clear gap in quality between real and synthetic data.}
    \label{fig:gen-vs-original}
\end{figure}

We evaluate how well models perform when trained exclusively on synthetic data generated through either prompting or the agentic workflow. Figure \ref{fig:gen-vs-original} shows that training on synthetic data alone leads to a large and consistent performance drop across all tasks, datasets, and model architectures. This confirms that current generation methods, despite improvement from agentic workflows, can not yet replace human-annotated ABSA data. Detailed per-task and per-dataset results for ATE, ATSC, and ASPE are provided in Tables \ref{tab:gen_ate}, \ref{tab:gen_atsc}, and \ref{tab:gen_aspe} in the Appendix.

A major contributor to this performance gap is the lack of linguistic richness in synthetic reviews. Real SemEval sentences often contain indirect sentiment, idioms, sarcasm, vague references, and domain-specific expressions that are difficult for generation models to reproduce. Even with stylistic conditioning, synthetic examples tend to be cleaner, more literal, and less ambiguous, reducing the diversity needed for robust generalization. In addition, as mentioned earlier (Table \ref{tab:label_cons}), synthetic data—especially from prompting—still contains label inconsistencies that introduce further noise. These semantic simplifications and residual labeling errors compound most severely in complex tasks like ASPE, widening the gap between synthetic-only and real-data training. 

\begin{table*}
\centering
\small
\setlength{\tabcolsep}{3.5pt}
\renewcommand{\arraystretch}{0.8} 
\begin{tabular}{llcccc}
\toprule
\textbf{Model} & \textbf{Data Type} & \textbf{Laptop14 } & \textbf{Rest14 } & \textbf{Rest15 } & \textbf{Rest16} \\
\midrule

\multirow{7}{*}{\textbf{T5-Base}} 
& Original Data            & 93.32     & 95.12     & 80.90     & 83.28    \\
\cmidrule(lr){2-6}
\cmidrule(lr){2-6}
\cmidrule(lr){2-6}
& \textbf{Mixed Data - Agentic} & & & & \\
& Mixed x1               & \textbf{93.82} \textcolor{gray}{↑0.50} & 94.63 \textcolor{gray}{↓0.49} & 80.98 \textcolor{gray}{↑0.08} & \textbf{85.21} \textcolor{gray}{↑1.93} \\
& Mixed x2               & 93.50 \textcolor{gray}{↑0.18} & 94.26 \textcolor{gray}{↓0.86} & \textbf{82.22} \textcolor{gray}{↑1.32} & 84.37 \textcolor{gray}{↑1.09} \\
\cmidrule(lr){2-6}
& \textbf{Mixed Data - Prompting} & & & & \\
& Mixed x1               & 92.83 \textcolor{gray}{↓0.49} & 94.53 \textcolor{gray}{↓0.59} & 80.43 \textcolor{gray}{↓0.47} & 83.83 \textcolor{gray}{↑0.55} \\
& Mixed x2               & 92.97 \textcolor{gray}{↓0.35} & 94.14 \textcolor{gray}{↓0.98} & 78.88 \textcolor{gray}{↓2.02} & 84.72 \textcolor{gray}{↑1.44} \\
\midrule

\multirow{7}{*}{\textbf{Tk-Instruct}} 
& Original Data            & 93.91     & 95.01     & 83.41     & 85.47     \\
\cmidrule(lr){2-6}
\cmidrule(lr){2-6}
& \textbf{Mixed Data - Agentic} & & & & \\
& Mixed x1               & 93.93 \textcolor{gray}{↑0.02} & 94.20 \textcolor{gray}{↓0.81} & \textbf{84.12} \textcolor{gray}{↑0.71} & \textbf{87.14} \textcolor{gray}{↑1.67} \\
& Mixed x2               & \textbf{94.07} \textcolor{gray}{↑0.16} & 93.34 \textcolor{gray}{↓1.67} & 82.50 \textcolor{gray}{↓0.91} & 86.38 \textcolor{gray}{↑0.91} \\
\cmidrule(lr){2-6}
& \textbf{Mixed Data - Prompting} & & & & \\
& Mixed x1               & 93.80 \textcolor{gray}{↓0.11} & 94.10 \textcolor{gray}{↓0.91} & 82.30 \textcolor{gray}{↓1.11} & 85.32 \textcolor{gray}{↓0.15} \\
& Mixed x2               & 93.35 \textcolor{gray}{↓0.56} & 94.20 \textcolor{gray}{↓0.81} & 80.00 \textcolor{gray}{↓3.41} & 84.39 \textcolor{gray}{↓1.08} \\
\bottomrule
\end{tabular}

\caption{ATE F\textsubscript{1} scores across datasets for each model trained on original data and mixed datasets combining original and synthetic examples. 
Mixed x1 and Mixed x2 denote synthetic-to-original data ratios of 1:1 and 2:1, respectively. 
Agentic augmentation provides clearer gains, while prompting-based augmentation yields limited or negative impact.}

\label{tab:mix_ate}
\end{table*}

\subsection{Training on Mixed Data} 
\label{sec:mix}

While synthetic data on its own does not match the performance of human-annotated training data, it can still have a positive impact when used in combination with real examples (Table \ref{tab:mix_ate} for the aspect term extraction task). In this setup, the synthetic data serves as a complementary source of variation. It helps reinforce the core task structure and boosts the diversity of examples the model sees, while the real data provides that essential anchor with linguistically rich and semantically consistent examples. Detailed per-task and per-dataset results for ATSC, and ASPE are provided in Tables \ref{tab:mix_atsc} and \ref{tab:mix_aspe} in the Appendix.

\begin{figure}
\centering
\includegraphics[width=0.47\textwidth]{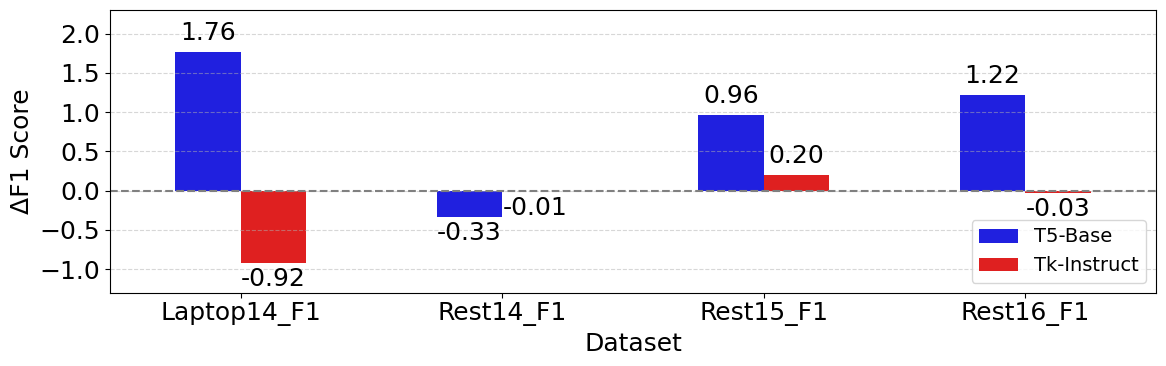}
\caption{ $\Delta$F\textsubscript{1} between baseline and added agentic data, across both \texttt{T5-Base} and \texttt{Tk-Instruct}. Each bar represents mean F\textsubscript{1} scores averaged over ATE, ATSC, and ASPE tasks for each dataset. This plot shows the clear difference of the augmentation effectiveness on these models.}
\label{fig:mix-vs-original}
\end{figure}

\subsection{Model Sensitivity to Data Augmentation: T5 vs. Tk-Instruct}
\label{sec:mix_model_compare}
Our results show that \texttt{T5-Base} and \texttt{Tk-Instruct} respond very differently to synthetic augmentation. Across tasks and datasets, \texttt{T5-Base} consistently benefits from agentic data in the Mixed x1 setting, with clear gains in ATSC and stable or slightly improved performance in ATE. In contrast, augmentation has only marginal effects on \texttt{Tk-Instruct}, and in some cases—particularly ASPE—slightly lowers performance. This pattern is also reflected in the aggregated task-level comparison (Figure~\ref{fig:mix-vs-original}), where \texttt{T5-Base} shows noticeably larger improvements from mixing real and synthetic data.

These differences can be attributed to the models’ pretraining. \texttt{Tk-Instruct} is heavily instruction-tuned on over 1,600 NLP tasks \cite{TK-INSTRUCT}, giving it broad exposure to task structures and linguistic patterns that already overlap with ABSA. As a result, its baseline performance is strong and additional synthetic examples add little new information, sometimes even introducing redundant or noisy patterns. \texttt{T5-Base}, lacking this instruction-driven prior, benefits much more from the additional variation and task-specific structure introduced by agentic augmentation.

\begin{figure}
\centering
\includegraphics[width=0.47\textwidth]{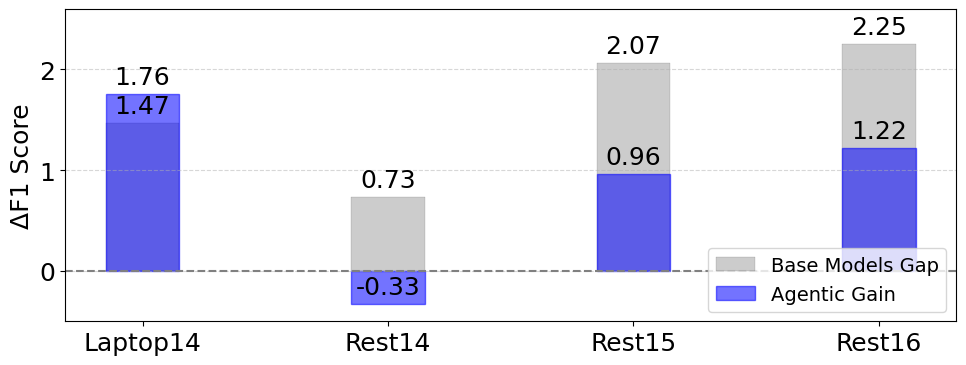} 
\caption{Agentic data augmentation narrows the F\textsubscript{1} gap between \texttt{T5-Base} and \texttt{Tk-Instruct}. Grey bars indicate the original model gap; blue bars show the performance gain of \texttt{T5-Base} after agentic augmentation (Mixed x1).}
\label{fig:agentic-gap-plot}
\end{figure}

This difference reinforces the idea that the utility of data augmentation is inversely related to the generalization capabilities already embedded in a model's pretraining. When a model has already been exposed to massive, diverse, and well-labeled task instructions, the benefits from new, generated examples are just not as significant.

Additionally, This difference has an important consequence: targeted synthetic augmentation can significantly narrow, and sometimes eliminate, the performance gap between the two models. Although \texttt{Tk-Instruct} consistently outperforms \texttt{T5-Base} when trained only on human-annotated data, the introduction of agentic synthetic examples (Mixed x1) leads to sharp improvements in \texttt{T5-Base}. As shown in Figure~\ref{fig:agentic-gap-plot}, these gains often bring \texttt{T5-Base} close to \texttt{Tk-Instruct} and even allow it to surpass the counterpart.

This shift highlights a key contribution of our work: task-specific agentic augmentation can substitute for large-scale instruction tuning. Whereas \texttt{Tk-Instruct} achieves its performance through massive, manually curated pretraining, our agentic pipeline is fully automatic and tailored directly to ABSA. Despite this simplicity, it provides substantial benefits to models like \texttt{T5-Base} and reduces their dependence on costly human-designed pretraining corpora.

\subsection{Agentic vs. Prompting-Based Augmentation}
\label{sec:mix_agent_vs_prompting}

\begin{figure}
\centering
\includegraphics[width=0.47\textwidth]{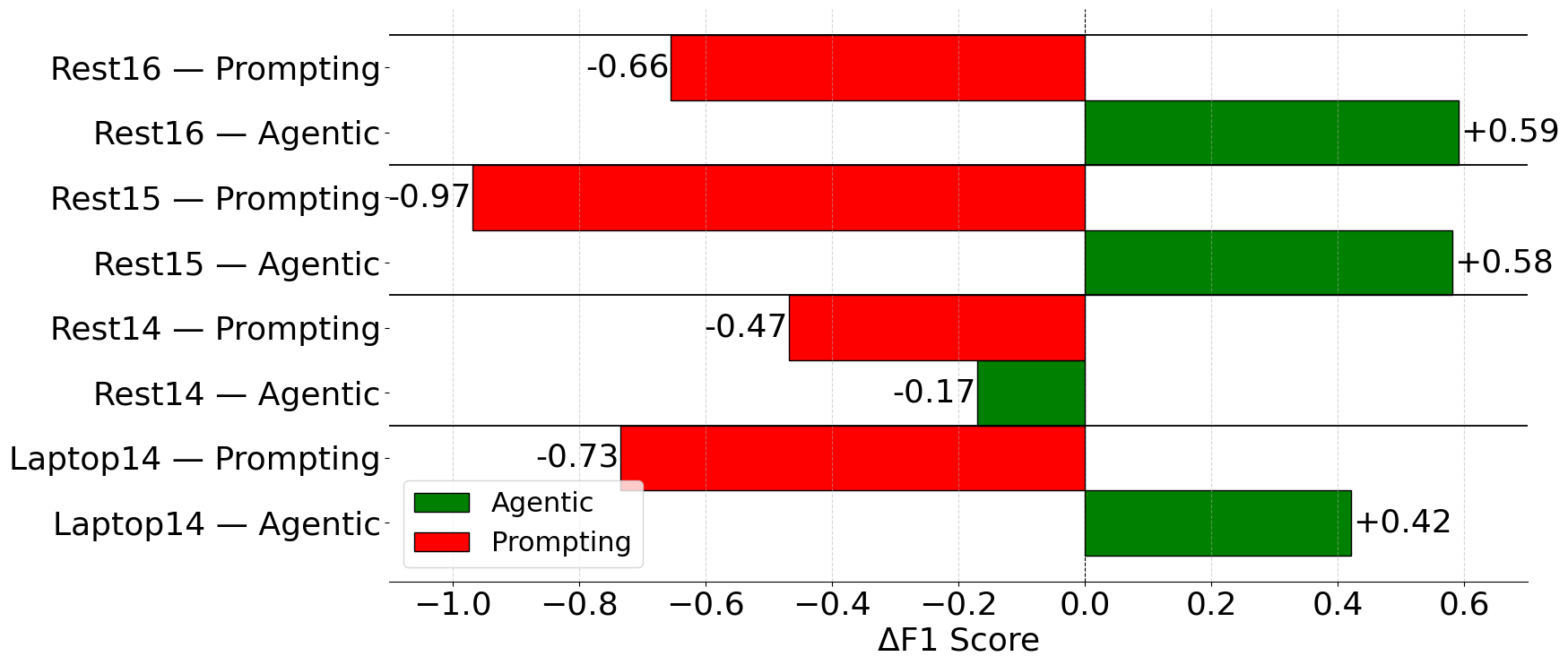}
\caption{Average F\textsubscript{1} score change ($\Delta F\textsubscript {1}$) for agentic and prompting augmentation, computed as the difference between the Mixed x1 setup and original-only baseline. Scores are averaged across three ABSA subtasks and two model architectures for each dataset.}
\label{fig:agent-vs-prompt}
\end{figure}

Since both augmentation methods use the same language model, the same prompts, and the same aspect–polarity sampling strategy, any difference in downstream performance is expected to come from the generation process itself. Across all tasks, datasets, and model architectures, agentic augmentation consistently improves or matches baseline performance, whereas prompting-based augmentation rarely helps and often degrades model accuracy. Figure~\ref{fig:agent-vs-prompt} summarizes this trend.

The advantage of the agentic workflow stems from its built-in evaluation and verification steps. Unlike prompting, which accepts every LLM output—even those with incorrect aspect terms, mismatched polarities, or overly simplistic phrasing—the agentic pipeline filters out noisy generations through explicit policy construction and label checking. As a result, agentic data achieves substantially higher label consistency (Table~\ref{tab:label_cons}), making it more beneficial when utilized either alone or mixed with real data.

\begin{figure}[t]
\centering
\includegraphics[width=0.5\textwidth]{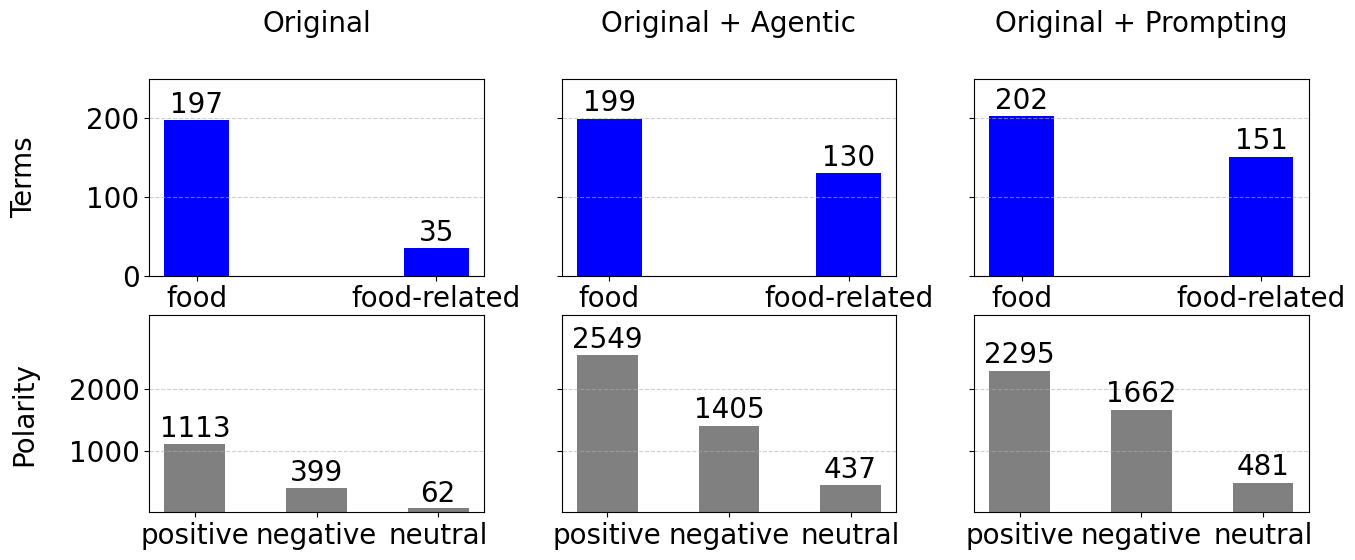}
\caption{Comparison of food-related term frequency and sentiment polarity in Rest 16 across original and mixed data sets, suggesting that augmentation mitigates the inherent biases.}
\label{fig:dist_compare}
\end{figure}

Both methods also employ uniform sampling to counteract label and term-frequency biases in the original datasets, and both help expose models to underrepresented aspect terms and polarities. Figure \ref{fig:dist_compare} depicts the distribution in SemEval 16 restaurant data set. However, only the agentic approach provides this benefit without introducing too much additional noise. Prompting-generated data suffers from low label accuracy \ref{tab:label_cons}, which cancels out the expected gains of bias correction and ultimately harms downstream training.

Overall, the evidence shows that high-quality augmentation requires more than good prompts: it requires structured generation, verification, and filtering. The agentic framework delivers this control, producing cleaner, more reliable synthetic examples, while naive prompting fails to meet the consistency that ABSA tasks demand.

\subsection{Task Sensitivity to Augmentation: ATE vs. ASPE vs. ATSC}
\label{sec:mix_task_compare}

\begin{figure}[t]
    \centering
    \includegraphics[width=0.45\textwidth]{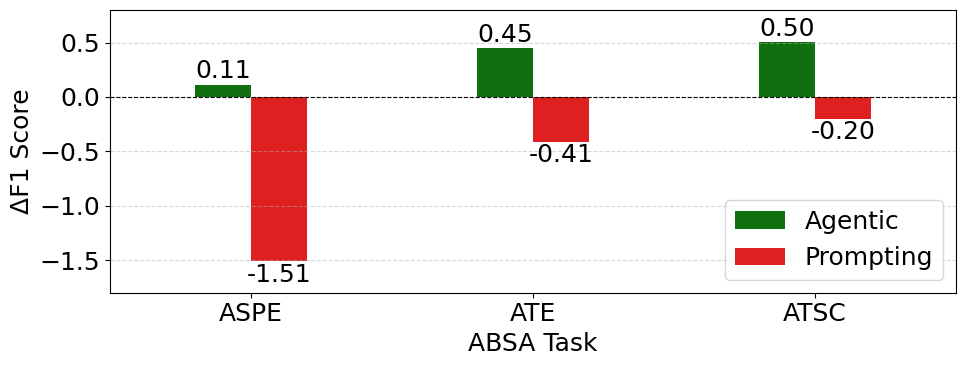}
    \caption{Average change in F\textsubscript{1} score across all datasets and both models for each ABSA task (ATE, ATSC, ASPE) under Mixed x1 conditions. This figure points out the effectiveness of augmentation for each task.}
    \label{fig:task_compare}
\end{figure}

The effects of data augmentation are not uniform across ABSA subtasks. As we can see in Figure \ref{fig:task_compare}, both ATE and ATSC benefit consistently from agentic synthetic data, showing clear improvements across most datasets and models. In contrast, gains in ASPE are limited and often inconsistent, especially for \texttt{Tk-Instruct}. This pattern aligns with the relative difficulty of the tasks: ATSC requires only sentiment classification for a given aspect, making it the easiest to support with synthetic examples. ATE is more challenging because the model must identify aspect spans, and ASPE is the most complex, requiring both extraction and sentiment assignment simultaneously.

The label consistency results in Table~\ref{tab:label_cons} reinforce this trend. Synthetic examples are highly reliable for ATSC, where the model only needs to express sentiment toward an already-given aspect. Because this task involves a single, explicit decision, both prompting and agentic generation tend to produce clean and well-aligned labels. For ATE, consistency drops: the model must identify aspect spans on its own, and generated sentences sometimes introduce ambiguity—such as multiword expressions, coordination, or slight wording mismatches—that make span extraction harder to learn from. ASPE shows the lowest consistency, as it compounds the challenges of both ATE and ATSC. Here, even small deviations (e.g., missing one term, splitting a phrase, misplacing a sentiment) break the aspect–sentiment pairing and introduce noise into training.

This downward progression in label reliability directly mirrors the downstream performance patterns. When the labels are simple and unambiguous, as in ATSC, augmentation provides strong and consistent improvements. When labels become harder to generate and validate, as in ATE and especially ASPE, the benefit diminishes and may even reverse. In essence, the harder the task, the more fragile augmentation becomes, because even subtle inconsistencies in synthetic labels can mislead the model and outweigh the value of increased data diversity.

\section{Conclusions and Future Work}

In this research, we explored the effectiveness of agentic data augmentation specifically for Aspect-Based Sentiment Analysis (ABSA). By comparing an agentic workflow to an otherwise identical prompting-based pipeline—using the same model, prompts, and sampling—we isolated the impact of structured generation and verification on downstream performance across three ABSA subtasks and four benchmark datasets. Our results show that, while synthetic data does not contain as much information as human annotations, high-quality agentic data improves performance when mixed with real examples. These gains are strongest for \texttt{T5-Base}, enabling it to approach or even surpass the baseline performance of the more extensively pretrained Tk-Instruct model. This demonstrates that task-specific, high-quality augmentation can partially close the gap created by large-scale instruction tuning.

We also find that agentic augmentation consistently outperforms naive prompting due to its higher label consistency and lower noise, despite using the same underlying model. Benefits are task-dependent: ATSC and ATE show clear improvements, while ASPE sees limited gains due to its higher structural complexity. Finally, augmentation is effective only in moderation—1:1 mixing yields improvements, but larger synthetic datasets dilute quality and reduce performance. Overall, these findings highlight that controlled, agentic generation provides a practical and scalable way to enhance ABSA models, offering meaningful gains without requiring extensive pretraining or additional human annotation.

Future work may extend this agentic augmentation framework to additional languages and domains, leveraging multilingual LLMs to broaden applicability with minimal structural changes. Another promising direction is to decompose the pipeline into task-specialized components—for example, using a strong text generator for sample creation and a dedicated ABSA model for validation—to further improve data quality. Incorporating adaptive sampling or confidence-based filtering could help prioritize informative and reliable synthetic examples, reducing noise during training. Finally, integrating model-in-the-loop feedback into the agentic workflow would allow the system to generate data that directly targets a model’s weaknesses, enabling more efficient, responsive, and context-aware augmentation.

\section{Limitations}
\label{sec:limitations}
This work is constrained by the capabilities of the open-source LLMs available to us. Although Qwen2.5 provides solid tool-calling performance, it remains less capable than proprietary state-of-the-art systems, limiting the complexity of the agent architecture we could implement. In addition, the selected model might be less capable than other larger open-source models, which can impact the quality of the decisions it makes during the process of data generation.

\section{Ethical Considerations}
This work uses publicly available SemEval datasets and open-source models such as T5, Tk-Instruct, and Qwen2.5. While widely adopted, these resources may contain biases or inaccuracies that can influence model behavior; a detailed analysis of such issues is beyond the scope of this study. We also acknowledge that sentiment analysis systems can be misused if deployed irresponsibly. Although our focus is strictly on technical aspects of data augmentation, we emphasize the need for careful, ethical application of these methods and encourage future work to examine fairness, bias, and potential downstream risks more closely.

\section*{Acknowledgments}
This work was supported by the BMFTR and the state of North Rhine-Westphalia as part of the Lamarr Institute for Machine Learning and Artificial Intelligence and the AISafety Project, funded under the BMFTR grant proposal No. 05D23PD1.

\bibliography{literature}

\section{Appendix}

\subsection{Agentic Generation and Evaluation Traces}

The following examples illustrate the agentic workflow for data generation and evaluation. The generation trace shows how sentences are produced under explicit stylistic and label constraints, while the evaluation trace demonstrates how incorrect or misaligned outputs are automatically filtered. Together, these examples highlight how the agentic pipeline enforces label correctness before synthetic data is accepted.

\onecolumn

\begin{tcolorbox}[title=Agentic Generation Trace Example, colback=gray!5, colframe=black!75, fonttitle=\bfseries]

\textbf{Agent Input:} \\
\texttt{generate a sentence with dataset information in mind.}

\vspace{0.3em}
\textbf{Agent Reasoning Trace:}
\begin{itemize}
  \item \textbf{Tool Call:} \texttt{get\_info()}
  \item \textbf{Tool Response:} \texttt{\{"writing\_style": "informal", "grammar\_structure": "simple sentences", "length": "medium"\}, "Terms": ['Ginger House', "Polarity": ['positive'] }

  \item \textbf{Tool Call:} \\ \texttt{generate\_sentences(style\_info= \{"writing\_style": "informal", "grammar\_structure": "simple sentences", "length": "medium"\})}
  \item \textbf{Tool Response:} \\
  \texttt{The Ginger House is a cozy spot that really warms the heart!} \\
  \texttt{Terms= ['Ginger House']} \\
  \texttt{Polarity= ['positive']}
\end{itemize}

\end{tcolorbox}

\begin{tcolorbox}[title=Agentic Evaluation Trace Example, colback=gray!5, colframe=black!75, fonttitle=\bfseries]

\textbf{Input:} \\
\texttt{The balcony was cramped and had limited tables, but I loved the view from it.}\\
\texttt{Terms= ['balcony'], Polarity= ['negative']}

\vspace{0.5em}
\textbf{Agent Trace:}
\begin{itemize}
  \item \textbf{Tool Call:} \texttt{label\_inclusion}
  \item \textbf{Tool Response:}  \texttt{OK}
  \item \textbf{Tool Call:} \texttt{evaluate\_sentence}
  \item \textbf{Tool Response:}  \texttt{NOT\_OK}
  \item \textbf{Final Decision:} \texttt{NOT\_OK} \\
  \small The generated sentence failed the second check
\end{itemize}

\end{tcolorbox}

\subsection{Additional Results}
\begin{table*}
\centering
\resizebox{ \textwidth}{!}{%
\begin{tabular}{|p{0.7cm}|p{7cm}|p{5cm}|p{5cm}|}
\hline
\textbf{ID} & \textbf{raw\_text} & \textbf{aspectTerms} & \textbf{aspectCategories} \\
\hline
2383 & Bottom line: B+ for the food, F for the service. & 
[\{term: \textit{food}, polarity: \textit{positive}\}, \{term: \textit{service}, polarity: \textit{negative}\}] & 
[\{category:\textit{food}, polarity: \textit{positive}\}, \{category: \textit{service}, polarity: \textit{negative}\}] \\
\hline
766 & We always go there on the weekends and leave extremely full and satisfied. & 
[\{term: \textit{noaspectterm}, polarity: \textit{none}\}] & 
[\{category: \textit{anecdotes}, polarity: \textit{positive}\}] \\
\hline
1419 & We actually gave 10\% tip (which we have never done despite mediocre food and service), because we felt totally ripped off. & 
[\{term: \textit{food}, polarity: \textit{neutral}\}, \{term: \textit{service}, polarity: \textit{neutral}\}, \{term: \textit{tip}, polarity: \textit{negative}\}] & 
[\{category: \textit{food}, polarity: \textit{neutral}\}, \{category: \textit{service}, polarity: \textit{neutral}\}, \{category: \textit{price}, polarity: \textit{negative}\}] \\
\hline
1700 & The food was good. & 
[\{term: \textit{food}, polarity: \textit{positive}\}] & 
[\{category: \textit{food}, polarity: \textit{positive}\}] \\
\hline
1892 & Service is great, takeout is good too. & 
[\{term: \textit{Service}, polarity: \textit{positive}\}, \{term: \textit{takeout}, polarity: \textit{positive}\}] & 
[\{category: \textit{food}, polarity: \textit{positive}\}, \{category: \textit{service}, polarity: \textit{positive}\}] \\
\hline
\end{tabular}
} 
\caption{Sample entries from SemEval-2014 Restaurant dataset}
\label{tab:absa-raw-format}
\end{table*}

\begin{table*}
\centering
\small
\setlength{\tabcolsep}{3.5pt}
\renewcommand{\arraystretch}{0.8} 
\begin{tabular}{llcccc}
\toprule
\textbf{Model} & \textbf{Data Type} & \textbf{Laptop14 } & \textbf{Rest14 } & \textbf{Rest15 } & \textbf{Rest16 } \\
\midrule
\multirow{7}{*}{\textbf{T5-Base}} 
& Original Data            & 93.32     & 95.12     & 80.90     & 83.28    \\
\cmidrule(lr){2-6}
& \textbf{Generated Data - Agentic} & & & & \\
& Gen. x1                 & 46.18  & 66.53  & 50.82  & 54.73  \\
& Gen. x2                 & 46.83  & 65.09  & 50.19  & 54.30  \\
\cmidrule(lr){2-6}
& \textbf{Generated Data - Prompting} & & & & \\
& Gen. x1                 & 41.46  & 58.85  & 48.45  & 51.13  \\
& Gen. x2                 & 40.35  & 55.73  & 46.49  & 49.05 \\
\midrule
\multirow{7}{*}{\textbf{Tk-Instruct}} 
& Original Data            & 93.91     & 95.01     & 83.41     & 85.47     \\
\cmidrule(lr){2-6}
& \textbf{Generated Data - Agentic} & & & & \\
& Gen. x1                 & 47.23  & 65.23  & 50.63  & 55.69  \\
& Gen. x2                 & 45.39  & 64.00  & 50.82  & 52.02  \\
\cmidrule(lr){2-6}
& \textbf{Generated Data - Prompting} & & & & \\
& Gen. x1                 & 39.69  & 55.38  & 44.67  & 45.86  \\
& Gen. x2                 & 39.26  & 51.71  & 43.89  & 43.83  \\
\bottomrule
\end{tabular}
\caption{ATE F\textsubscript{1} scores across datasets for each model trained on original data and on synthetic data only. Gen.\ x1 and Gen.\ x2 indicate synthetic data generated at 1:1 and 2:1 ratios relative to the original training set. Agentic generation consistently outperforms prompting, while synthetic-only training shows lower performance.}
\label{tab:gen_ate}
\end{table*}

\begin{table*}
\centering
\small
\setlength{\tabcolsep}{3.5pt}
\renewcommand{\arraystretch}{0.8} 
\begin{tabular}{llcccc}
\toprule
\textbf{Model} & \textbf{Data Type} & \textbf{Laptop14 } & \textbf{Rest14 } & \textbf{Rest15 } & \textbf{Rest16 } \\
\midrule

\multirow{7}{*}{\textbf{T5-Base}} 
& Original Data            & 79.78    & 86.25     & 88.74     & 91.86    \\
\cmidrule(lr){2-6}
& \textbf{Generated Data - Agentic} & & & & \\
& Gen. x1                 & 77.42  & 79.64  & 81.54  & 88.29  \\
& Gen. x2                 & 78.05  & 80.00  & 81.73  & 89.26  \\
\cmidrule(lr){2-6}
& \textbf{Generated Data - Prompting} & & & & \\
& Gen. x1                 & 75.86  & 76.85  & 81.84  & 87.57  \\
& Gen. x2                 & 74.92  & 75.53  & 80.44  & 86.66 \\
\midrule
\multirow{7}{*}{\textbf{Tk-Instruct}} 
& Original Data           & 82.75 & 87.41 & 88.56 & 94.30 \\
\cmidrule(lr){2-6}
& \textbf{Generated Data - Agentic} & & & & \\
& Gen. x1                 & 74.60 & 79.37 & 80.44 & 88.78 \\
& Gen. x2                 & 76.64 & 79.10 & 79.88 & 89.26 \\
\cmidrule(lr){2-6}
& \textbf{Generated Data - Prompting} & & & & \\
& Gen. x1                 & 73.04 & 76.25 & 82.28 & 86.17 \\
& Gen. x2                 & 72.57 & 76.25 & 79.70 & 85.04 \\

\bottomrule
\end{tabular}
\caption{ATSC F\textsubscript{1} scores across datasets for each model trained on original data and on synthetic data only. 
Gen.\ x1 and Gen.\ x2 indicate synthetic data generated at 1:1 and 2:1 ratios relative to the original training set. 
Agentic generation consistently outperforms prompting, while synthetic-only training shows lower performance.}

\label{tab:gen_atsc}
\end{table*}

\begin{table*}
\centering
\small
\setlength{\tabcolsep}{3.5pt}
\renewcommand{\arraystretch}{0.8} 
\begin{tabular}{llcccc}
\toprule
\textbf{Model} & \textbf{Data Type} & \textbf{Laptop14 } & \textbf{Rest14 } & \textbf{Rest15 } & \textbf{Rest16 } \\
\midrule

\multirow{7}{*}{\textbf{T5-Base}} 
& Original Data            & 81.19 & 81.34 & 73.38 & 79.47 \\
\cmidrule(lr){2-6}
& \textbf{Generated Data - Agentic} & & & & \\
& Gen. x1                 & 28.42 & 48.79 & 37.94 & 41.49 \\
& Gen. x2                 & 28.81 & 48.88 & 37.58 & 42.56 \\
\cmidrule(lr){2-6}
& \textbf{Generated Data - Prompting} & & & & \\
& Gen. x1                 & 23.13 & 38.15 & 28.58 & 35.30 \\
& Gen. x2                 & 22.80 & 36.00 & 29.09 & 34.51 \\
\midrule
\multirow{7}{*}{\textbf{Tk-Instruct}} 
& Original Data           & 82.03 & 82.49 & 77.25 & 81.60 \\
\cmidrule(lr){2-6}
& \textbf{Generated Data - Agentic} & & & & \\
& Gen. x1                 & 31.98 & 50.07 & 40.02 & 46.63 \\
& Gen. x2                 & 31.72 & 50.60 & 41.04 & 46.83 \\
\cmidrule(lr){2-6}
& \textbf{Generated Data - Prompting} & & & & \\
& Gen. x1                 & 23.47 & 39.47 & 31.87 & 33.12 \\
& Gen. x2                 & 22.04 & 36.03 & 31.32 & 33.12 \\

\bottomrule
\end{tabular}
\caption{ASPE F\textsubscript{1} scores across datasets for each model when trained on original data and on synthetic data only. 
Gen.\ x1 and Gen.\ x2 denote synthetic datasets generated at ratios of 1:1 and 2:1 relative to the original training set size. 
The performance gap between synthetic-only and original data is most pronounced for ASPE, highlighting the difficulty of learning joint aspect–sentiment extraction from generated data alone, even with agentic generation.}

\label{tab:gen_aspe}
\end{table*}

\vspace{2cm}

\begin{table*}
\centering
\small
\setlength{\tabcolsep}{3.5pt}
\renewcommand{\arraystretch}{0.8} 
\begin{tabular}{llcccc}
\toprule
\textbf{Model} & \textbf{Data Type} & \textbf{Laptop14 } & \textbf{Rest14 } & \textbf{Rest15 } & \textbf{Rest16} \\
\midrule

\multirow{7}{*}{\textbf{T5-Base}} 
& Original Data            & 79.78    & 86.25     & 88.74     & 91.86    \\
\cmidrule(lr){2-6}
\cmidrule(lr){2-6}
& \textbf{Mixed Data - Agentic} & & & & \\
& Mixed x1               & \textbf{83.38} \textcolor{gray}{↑3.60} & \textbf{86.78} \textcolor{gray}{↑0.53} & \textbf{88.92} \textcolor{gray}{↑0.18} & \textbf{93.33} \textcolor{gray}{↑1.47} \\
& Mixed x2               & 81.34 \textcolor{gray}{↑1.56} & 86.42 \textcolor{gray}{↑0.17} & 88.56 \textcolor{gray}{↓0.18} & 93.17 \textcolor{gray}{↑1.31} \\
\cmidrule(lr){2-6}
& \textbf{Mixed Data - Prompting} & & & & \\
& Mixed x1               & 79.93 \textcolor{gray}{↑0.15} & 86.96 \textcolor{gray}{↑0.71} & 88.37 \textcolor{gray}{↓0.37} & 93.82 \textcolor{gray}{↑1.96} \\
& Mixed x2               & 81.34 \textcolor{gray}{↑1.56} & 85.62 \textcolor{gray}{↓0.63} & 88.37 \textcolor{gray}{↓0.37} & 93.00 \textcolor{gray}{↑1.14} \\
\midrule

\multirow{7}{*}{\textbf{Tk-Instruct}} 
& Original Data           & 82.75 & 87.41 & 88.56 & 94.30 \\
\cmidrule(lr){2-6}
\cmidrule(lr){2-6}
& \textbf{Mixed Data - Agentic} & & & & \\
& Mixed x1               & 80.72 \textcolor{gray}{↓2.03} & \textbf{88.30} \textcolor{gray}{↑0.89} & \textbf{88.92} \textcolor{gray}{↑0.36} & 93.33 \textcolor{gray}{↓0.97} \\
& Mixed x2               & 81.34 \textcolor{gray}{↓1.41} & 87.41 \textcolor{gray}{↓0.00} & 88.74 \textcolor{gray}{↑0.18} & 92.68 \textcolor{gray}{↓1.62} \\
\cmidrule(lr){2-6}
& \textbf{Mixed Data - Prompting} & & & & \\
& Mixed x1               & 80.87 \textcolor{gray}{↓1.88} & 87.58 \textcolor{gray}{↑0.17} & 88.00 \textcolor{gray}{↓0.56} & 92.52 \textcolor{gray}{↓1.78} \\
& Mixed x2               & 81.19 \textcolor{gray}{↓1.56} & 87.23 \textcolor{gray}{↓0.18} & 87.60 \textcolor{gray}{↓0.96} & 92.84 \textcolor{gray}{↓1.46} \\
\bottomrule
\end{tabular}
\caption{ATSC F\textsubscript{1} scores across datasets for each model trained on original data and mixed datasets combining original and synthetic examples. 
Mixed x1 and Mixed x2 denote synthetic-to-original data ratios of 1:1 and 2:1; 
Agentic augmentation yields the largest and most consistent gains for T5-Base, whereas Tk-Instruct shows mixed or negative changes under both augmentation methods.}

\label{tab:mix_atsc}
\end{table*}

\begin{table*}
\centering
\small
\setlength{\tabcolsep}{3.5pt}
\renewcommand{\arraystretch}{0.8} 
\begin{tabular}{llcccc}
\toprule
\textbf{Model} & \textbf{Data Type} & \textbf{Laptop14 } & \textbf{Rest14 } & \textbf{Rest15 } & \textbf{Rest16 } \\
\midrule

\multirow{7}{*}{\textbf{T5-Base}} 
& Original Data            & 81.19 & 81.34 & 73.38 & 79.47 \\
\cmidrule(lr){2-6}
\cmidrule(lr){2-6}
\cmidrule(lr){2-6}
& \textbf{Mixed Data - Agentic} & & & & \\
& Mixed x1               & \textbf{82.37} \textcolor{gray}{↑1.18} & 80.31 \textcolor{gray}{↓1.03} & \textbf{76.00} \textcolor{gray}{↑2.62} & \textbf{79.72} \textcolor{gray}{↑0.25} \\
& Mixed x2               & 78.84 \textcolor{gray}{↓2.35} & \textbf{81.52} \textcolor{gray}{↑0.18} & 73.89 \textcolor{gray}{↑0.51} & 78.79 \textcolor{gray}{↓0.68} \\
\cmidrule(lr){2-6}
& \textbf{Mixed Data - Prompting} & & & & \\
& Mixed x1               & 80.68 \textcolor{gray}{↓0.51} & 80.42 \textcolor{gray}{↓0.92} & 72.92 \textcolor{gray}{↓0.46} & 79.02 \textcolor{gray}{↓0.45} \\
& Mixed x2               & 78.71 \textcolor{gray}{↓2.48} & 80.42 \textcolor{gray}{↓0.92} & 72.34 \textcolor{gray}{↓1.04} & 76.06 \textcolor{gray}{↓3.41} \\
\midrule

\multirow{7}{*}{\textbf{Tk-Instruct}} 
& Original Data           & 82.03 & 82.49 & 77.25 & 81.60 \\
\cmidrule(lr){2-6}
\cmidrule(lr){2-6}
& \textbf{Mixed Data - Agentic} & & & & \\
& Mixed x1               & 81.29 \textcolor{gray}{↓0.74} & 82.38 \textcolor{gray}{↓0.11} & 76.79 \textcolor{gray}{↓0.46} & 80.80 \textcolor{gray}{↓0.80} \\
& Mixed x2               & 80.62 \textcolor{gray}{↓1.41} & 81.66 \textcolor{gray}{↓0.83} & 75.60 \textcolor{gray}{↓1.65} & \textbf{82.21} \textcolor{gray}{↑0.61} \\
\cmidrule(lr){2-6}
& \textbf{Mixed Data - Prompting} & & & & \\
& Mixed x1               & 80.46 \textcolor{gray}{↓1.57} & 81.22 \textcolor{gray}{↓1.27} & 74.41 \textcolor{gray}{↓2.84} & 77.54 \textcolor{gray}{↓4.06} \\
& Mixed x2               & 79.70 \textcolor{gray}{↓2.33} & 81.52 \textcolor{gray}{↓0.97} & 74.10 \textcolor{gray}{↓3.15} & 81.30 \textcolor{gray}{↓0.30} \\
\bottomrule
\end{tabular}
\caption{ASPE F\textsubscript{1} scores across datasets for each model trained on original data and mixed datasets combining original and synthetic examples. 
Mixed x1 and Mixed x2 denote synthetic-to-original data ratios of 1:1 and 2:1, respectively; Agentic augmentation yields modest and dataset-dependent improvements. Prompting-based augmentation generally leads to performance drops for both models.}

\label{tab:mix_aspe}
\end{table*}

\end{document}